\documentclass[10pt,twocolumn,letterpaper]{article}

\usepackage{cvpr}
\usepackage{times}
\usepackage{epsfig}
\usepackage{graphicx}
\usepackage{amsmath}
\usepackage{amssymb}
\usepackage{siunitx}
\usepackage{bm}
\usepackage{subfig}

\usepackage[pagebackref=true,breaklinks=true,letterpaper=true,colorlinks,bookmarks=false]{hyperref}

 \cvprfinalcopy % *** Uncomment this line for the final submission

\def\cvprPaperID{****} % *** Enter the CVPR Paper ID here

\ifcvprfinal\fi
\begin{document}

%%%%%%%%% TITLE
\title{Learning Preference-Based Similarities from Face Images using Siamese Multi-Task CNNs}

\author{Nils Gessert \quad \quad \quad \quad \quad Alexander Schlaefer\\
\small Institute of Medical Technology, Hamburg University of Technology\\
%\small Am Schwarzenberg-Campus 3, 21073 Hamburg, Germany\\
{\tt\small nils.gessert@tuhh.de}
%\
}

\maketitle
%\thispagestyle{empty}

%%%%%%%%% ABSTRACT
\begin{abstract}
   Online dating has become a common occurrence over the last few decades. A key challenge for online dating platforms is to determine suitable matches for their users. A lot of dating services rely on self-reported user traits and preferences for matching. At the same time, some services largely rely on user images and thus initial visual preference. Especially for the latter approach, previous research has attempted to capture users' visual preferences for automatic match recommendation. These approaches are mostly based on the assumption that physical attraction is the key factor for relationship formation and personal preferences, interests, and attitude are largely neglected. Deep learning approaches have shown that a variety of properties can be predicted from human faces to some degree, including age, health and even personality traits. Therefore, we investigate the feasibility of bridging image-based matching and matching with personal interests, preferences, and attitude. We approach the problem in a supervised manner by predicting similarity scores between two users based on images of their faces only. The ground-truth for the similarity matching scores is determined by a test that aims to capture users' preferences, interests, and attitude that are relevant for forming romantic relationships. The images are processed by a Siamese Multi-Task deep learning architecture. We find a statistically significant correlation between predicted and target similarity scores. Thus, our results indicate that learning similarities in terms of interests, preferences, and attitude from face images appears to be feasible to some degree. %For binary classification into positive and negative matches we achieve an accuracy of $\SI{70}{\percent}$.
   
\end{abstract}

%%%%%%%%% BODY TEXT
\section{Introduction}

Online dating has taken an important role in today's society as more and more romantic relationships emerge from the use of online dating services~\cite{smith2013online}. In general, online dating services provide matches for their users that are likely to end in a romantic encounter or relationship. Some online dating services rely on self-reported personality traits, preferences and values that are obtained by requiring users to fill out a questionnaire. The results of this questionnaire are fed into an algorithm that determines suitable partners. Typically, dating services claim that their algorithms allow for uniquely tailored match selection for each individual user with a high probability of a positive first impression upon meeting in person or even long-term relationship success \cite{finkel2012online}. While such services do not always provide empirical evidence for their claims, research has suggested that personality traits and shared preferences and values have an important role in relationship satisfaction and longevity \cite{campbell2016initial,mcnulty2016should}. At the same time, services such as Tinder have emerged where partner selection is largely based on initial visual impressions and preferences. While these types of services generally target a different audience, initial visual impressions are a key factor for attraction between people \cite{willis2006first}. For automatic match recommendation with images, a machine learning method has been proposed where both visual features and previous user ratings are considered \cite{rothe2016some}. Similar to recommendation approaches using personality-based features \cite{krzywicki2010interaction}, a collaborative filtering method was used. So far, image-based match recommendation is rare and has largely been focused on physical attractiveness. This constitutes a gap to traditional matching approaches using, e.g., interests, preferences and personality-based properties. 

%Therefore, research has been conducted on predicting physical attractiveness from face images~\cite{whitehill2008personalized,gray2010predicting,wang2014attractive,liang2018scut}.

Face images have been used to derive a variety of properties using automated image analysis. For example, demographic properties such as age \cite{fu2010age} and sex \cite{ng2012recognizing} have been derived from faces. In particular, deep learning methods have been successful for these tasks \cite{levi2015age}. Also, predicting more subjective quantities such as facial beauty has been studied with machine learning methods \cite{yan2014cost,gan2014deep}. Even more subtle attributes such as self-reported personality traits have been investigated in terms of their predictability from face images with deep learning methods \cite{zhang2017physiognomy,gavrilescu2017predicting}. Thus, a variety of properties have been automatically derived from face images. This opens up the question of whether features that are particularly relevant for matching, besides facial beauty and attractiveness, can be predicted from face images.

In this work, we study the idea of predicting two people's similarity in terms of interests, preferences, and attitude based on face images for match recommendation. We feed two face images of two people into a Siamese multi-task deep learning model and predict how well the two users match in terms of different similarity criteria. These criteria include more superficial aspects such as general interests and leisure activities and also criteria with more depth such as lifestyle, relationship preferences, and value systems. 

Intuitively, similarity in terms of properties such as interests and lifestyle is very hard to predict from faces alone. However, lifestyle and leisure activity choices could affect facial properties over time and provide a rough indication of similarity between two people. At the same time, pose or facial expression might also provide information as previous research demonstrated that the choice of pictures uploaded to social media profiles is linked to personality \cite{liu2016analyzing}. Thus, our approach can provide interesting insights on the predictability of important similarity criteria for romantic relationships and, at the same time, also provide the advantage of convenience for dating platforms.

To study this challenging task, a dataset consisting of \num{6000} users was collected for model training and evaluation. For each user, one or several images showing the person's face are available. Also, each user took an initial test where interests, preferences, and attitude were reported. Based on these results, similarity matching scores were calculated between every user in the dataset. The final matching scores and images were kindly provided to use by the dating platform LemonSwan.

We address the proposed supervised learning problem with a DenseNet-based \cite{huang2017densely} convolutional neural network (CNN) architecture that is based on Siamese CNN architectures \cite{yi2014deep} and multi-task relationship learning concepts \cite{caruana1997multitask}. After an initial Siamese processing path, we jointly process image features in individual output paths for the different similarity matching scores. 

The main contributions of this paper are threefold. First, we demonstrate that directly learning preference-based similarity matching scores for dating recommendations from face images appears to be feasible. Second, we design a Siamese multi-task CNN architecture and analyze its properties for this particular learning problem. Third, we provide insights into the predictability and characteristics of different similarity matching scores from face images.

\section{Related Work}

The problem at hand is related to predicting properties from face images, automatic match recommendation systems and deep learning methods for image matching tasks.

\textbf{Predicting properties from faces} with automated methods has been widely studied. More objective, demographic properties such as age and sex have been predicted from face images using both conventional machine learning methods \cite{fu2010age,ng2012recognizing} and deep learning methods \cite{levi2015age,niu2016ordinal,rothe2018deep}. More subjective properties such as attractiveness and facial beauty are more difficult to quantify but can be approximated by averaging multiple ratings \cite{eisenthal2006facial,altwaijry2013relative,yan2014cost,liang2018scut}. Gan et al. proposed to use deep learning methods for facial beauty estimation \cite{gan2014deep} which was extended by several approaches with cascaded training \cite{xu2017facial} or multi-task learning \cite{gao2018automatic}. Besides facial beauty, health-related properties have also been predicted from faces using deep learning. For example, facial cues for congenital and neurodevelopmental disorders have been predicted with CNNs \cite{gurovich2019identifying}. Similarly, liftstyle-related properties in terms of the body mass index can be derived from faces \cite{barr2018detecting}.

In addition, less superficial properties such as personality traits have also been studied in terms of their predictability from face images. Predicting personality from faces is a common task that people unconsciously perform daily in social interactions. Thus, research has been conducted on human performance for personality trait prediction from faces. Earlier work studied how well self-reported personality traits match the assessment of strangers, based on short encounters with significant correlations between $r \approx 0.2$ to $r \approx 0.4$ \cite{albright1988consensus,watson1989strangers}. Similarly, other studies have found a weak but significant correlation between self-reported and predicted personality traits, based on photographs \cite{zebrowitz1997accurate,hassin2000facing,penton2006personality}. Although personality traits appear to be somewhat predictable, there can be variations in predictions for the same person and different face images, e.g., due to facial expression and pose \cite{todorov2008understanding,naumann2009personality,zebrowitz2011ecological,todorov2015social}. For automated recognition of personality traits, two deep learning methods have been presented where accuracies between $\SI{50}{\percent}$ and $\SI{80}{\percent}$ were reported for different personality traits \cite{zhang2017physiognomy,gavrilescu2017predicting}, measured by the 16 Personality Factors model. Also, a patent was filed recently for automatic personality and capability prediction from faces \cite{wilf2015method}.

Summarized, a plethora of properties can be partially predicted from face images. The properties studied in this work, i.e. interests, preferences, and attitude are similar to lifestyle- and personality-based properties, however, they have not been explicitly studied, in particular not in the context of similarity between two people.

\textbf{Automatic match recommendation} is a challenging task as, in contrast to typical product recommendations, the result also needs to be satisfactory for the recipient \cite{kutty2014people}. Often, recommendation services first collect user attributes such as preferences and education \cite{whyte2017things} or self-reported personality traits \cite{fiore2010s,clemens2015influence} which are then used in a match prediction algorithm. Some fully automated approaches have relied on content-based or collaborative filtering with user ratings of dating profiles \cite{brozovsky2007recommender,cai2010learning,krzywicki2010interaction,akehurst2011ccr}. Another approach relied on text messages exchanged between users as a feature basis for a machine learning model that predicts matches \cite{tu2014online}. This has been extended by engineering similarity scores based on mutual interests, initial messages and their responses \cite{xia2016design}. In addition, features for facial attractiveness have been integrated into the feature engineering process \cite{li2019incorporating}. Also, facial features have been used in a framework, where user ratings of face images are integrated into a collaborative filtering method \cite{rothe2016some}. 

\textbf{Deep learning methods} have been successful for a variety of supervised learning tasks such as image classification \cite{krizhevsky2012imagenet} and object detection \cite{girshick2015fast}. Learning tasks where two images need to processed simultaneously include image retrieval \cite{qi2016sketch} and person reidentification \cite{chung2017two}. For this type of task, Siamese CNNs are usually employed \cite{yi2014deep}. Here, CNNs process each image and predict a score that is used to calculate a similarity metric for learning \cite{hadsell2006dimensionality}. The learning target for the task at hand is different from problems such as person re-identification as we know the distance between pairs (matching scores). Still, we can reuse the concept of Siamese CNNs for building our deep learning model. In terms of the model output, the task at hand can also be seen as a multi-task learning (MTL) problem as we aim to predict different types of similarity scores. MTL approaches generally perform shared processing first, followed by different prediction strategies for the different tasks \cite{ruder2017overview}. In the case of hard parameter sharing, all tasks share the same parameters up to the model output \cite{caruana1997multitask}. For soft parameter sharing, the different tasks learn individual parameters that can be additionally constrained to enforce similarity for similar tasks \cite{misra2016cross}. MTL methods have been applied to a variety of deep learning methods, including face-based landmark detection with simultaneuous age regression, gender classification, pose estimation and facial attribute recognition \cite{zhang2014facial,ranjan2017hyperface}.

\section{Methods}

\begin{figure}[t]
\begin{center}
   \includegraphics[width=0.9\linewidth]{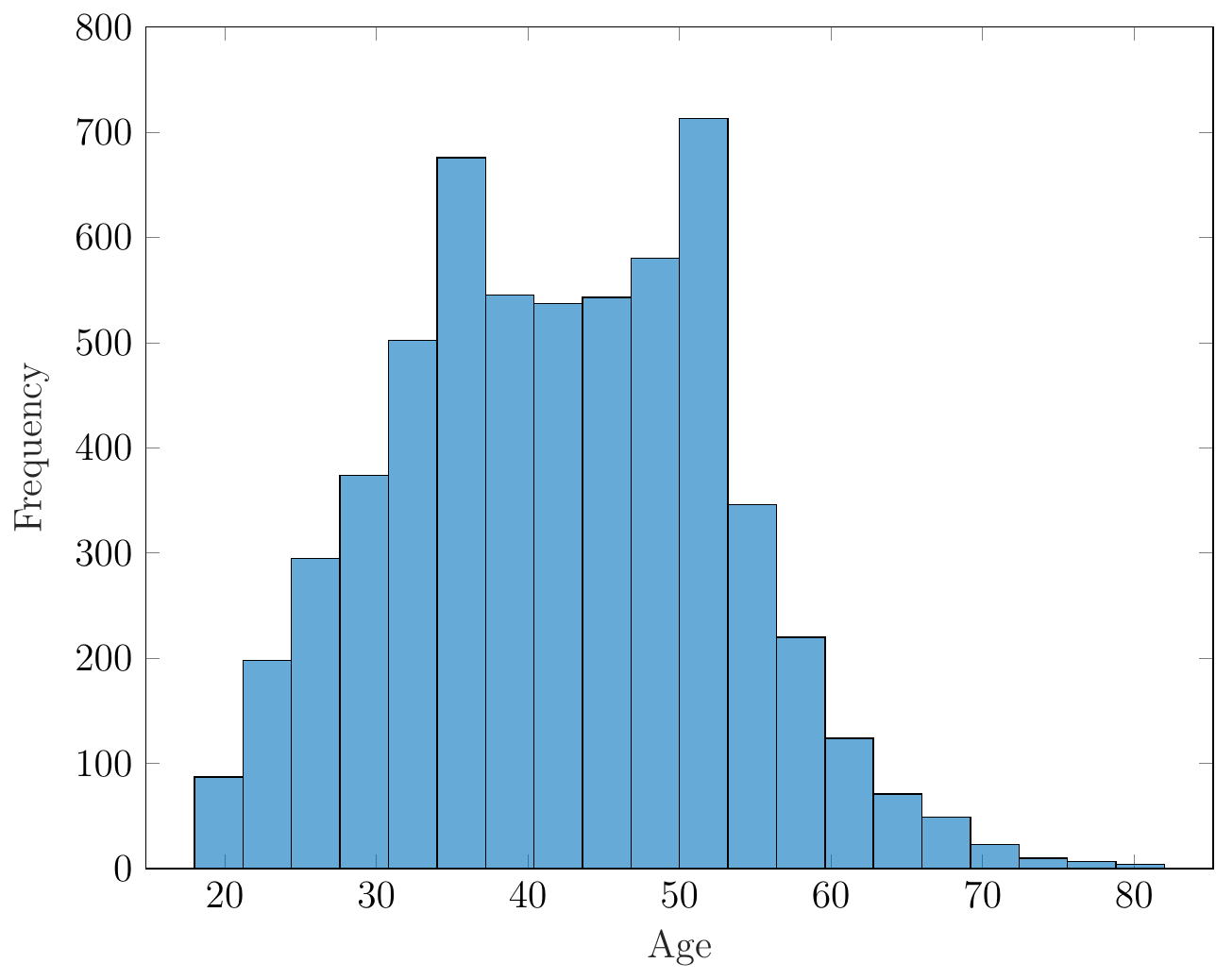}
\end{center}
   \caption{Age distribution in our dataset.}
\label{fig:age_dist}
\end{figure}

\subsection{Dataset}

All the data we use in this study was kindly provided to us by LemonSwan. The dataset we use contains $6000$ users with $8300$ face images. The images vary in quality and were handpicked from a larger pool such that low-quality images and images with multiple or very small faces were removed. The mean age of users is $42 \pm 11$ years and $\SI{51}{\percent}$ of users are female. The age distribution is shown in Figure~\ref{fig:age_dist}. Note that the age is not uniformly distributed and there is fewer user data available for very young and very old age groups. The users are largely Caucasian and most users live in Germany, the primary market of LemonSwan.

\begin{figure}[t]
\begin{center}
   \includegraphics[width=0.9\linewidth]{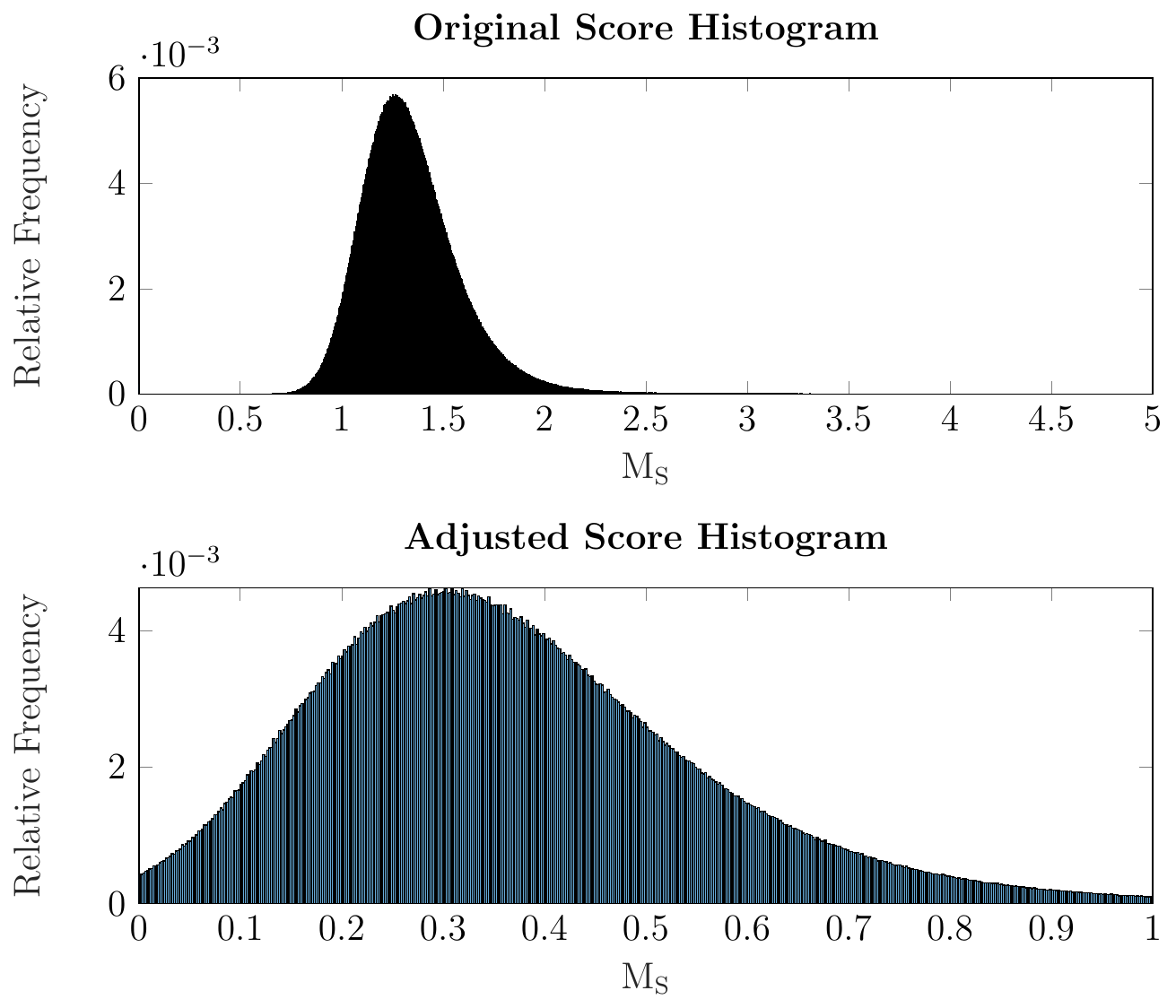}
\end{center}
   \caption{Original score histogram in our dataset (top) and the adjusted score histogram (bottom). For the adjusted score, we map the 1\textsuperscript{st} and 99\textsuperscript{th} percentile values to $0$ and $1$, respectively.}
\label{fig:scores}
\end{figure}

\begin{figure*}
\begin{center}
   \includegraphics[width=1.0\linewidth]{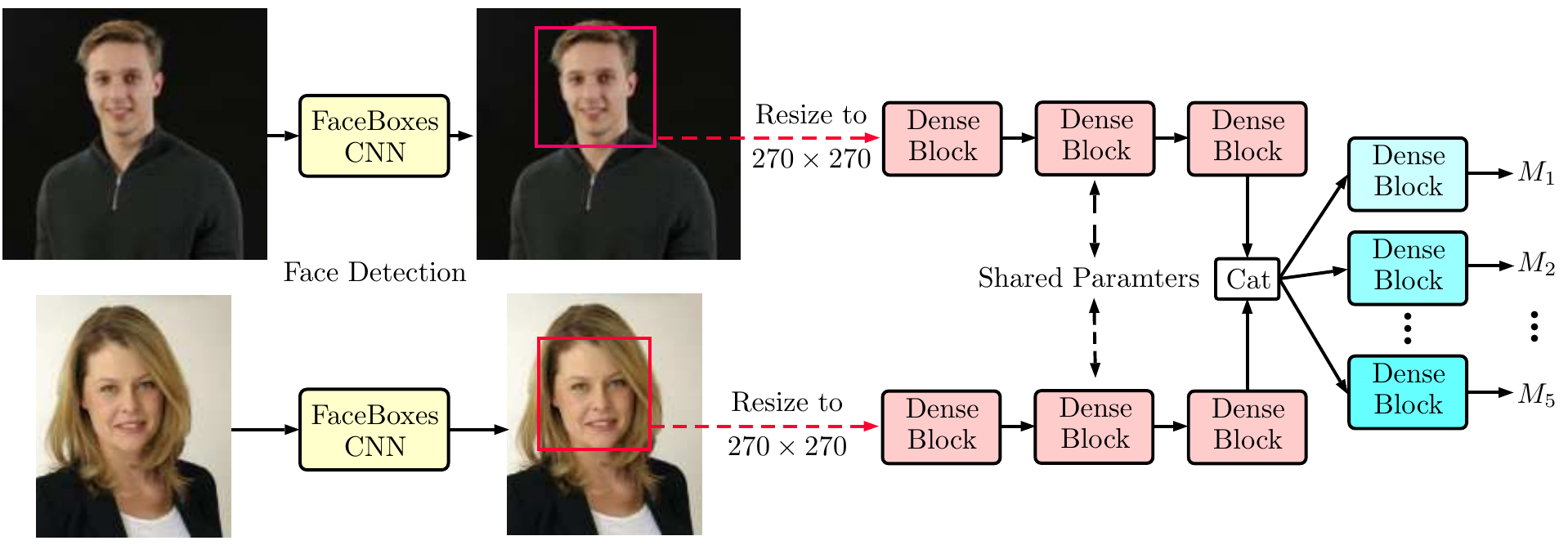}
\end{center}
   \caption{The processing pipeline we propose. FaceBoxes \cite{zhang2017faceboxes} is a light-weight CNN for face detection. In the matching CNN, the blue Dense Blocks do not share parameters. We omitted individual convolutional layers and transition layers for clarity, see \cite{huang2017densely}.}
\label{fig:pipeline}
\end{figure*}

Every user conducted a test which is used to calculate matching scores between all users. The test covers several areas related to interests and preferences, attitude and mindset, and personality. In this work we focus on five criteria for matching, namely "leisure activities" ($M_1$), "general interests" ($M_2$), "relationship preferences" ($M_3$), "lifestyle" ($M_4$) and "value system" ($M_5$). We also consider an overall matching score $M_{\mathit{S}}$ which is the average of all individual criteria. The user answers several questions related to each criterion and afterward, a distance-based matching score is calculated for each criterion between potentially matching users. Each score takes a value between $0$ and $5$ where a lower value indicates a better match. The main rationale behind the different criteria is to establish a basis for relationship formation without a focus on long-term relationship success. Thus, high matches in terms of these criteria should form a basis for initial interest and communication due to similar interests and mindsets. Note that there are more criteria that we omitted for simplicity. The matching system was developed at LemonSwan, based on work by Bak \cite{bak2015gast}. 
The criteria we cover have different levels of depth, i.e., some are more superficial preferences ($M_1$ and $M_2$) while others go more into depth ($M_3$, $M_5$) and are closer linked to identity and personality.

Note that the actual values in our dataset are not uniformly distributed within the original range of $0$ to $5$. Therefore, we map all scores to the actual range using the 1\textsuperscript{st} and 99\textsuperscript{th} percentile for the values $0$ and $1$, see Figure~\ref{fig:scores}. We perform this step in order to capture most matching scores in an interpretable range. Also, we invert the scores in order to obtain a more intuitive representation where $0$ indicates a bad match and $1$ indicates a good match.

We sample a subset of matches from the distribution of all possible matches shown in Figure~\ref{fig:scores}. In particular, for training, we sample more matches from the distribution's tails where similarities are high or low and thus potentially more expressive than a large amount of matching scores around the distribution's median. Also, we partially train on matches between the same sex to increase robustness. For validation and testing, we sample matches from the full distribution. We partition the dataset into training, validation and test splits, depending on the experiment. All sets are completely disjoint, i.e., we only consider matches among users of the same set. For the test set, we split off \num{1000} users. In terms of matches, we consider a relevant subset in terms of age. For females, we consider an age range of $-5$ and $+10$ around the self-reported age. Similarly, the range for males is $-10$ and $+5$. These age ranges are based on observed preference among users by LemonSwan. We only evaluate heterosexual matches. Results are reported for this test set, unless indicated otherwise. We perform hyperparameter tuning on a smaller validation set of \num{500} users. %We also evaluate consistency with cross-validation experiments were we chose several partitions in a similar fashion. %During training, we allow for a broader range of matches between users.  

\subsection{Deep Learning Method}

Our deep learning pipeline is shown in Figure~\ref{fig:pipeline}. The first step is to perform face detection and cropping. We rely on the FaceBoxes \cite{zhang2017faceboxes} method which is designed in particular for fast execution on a CPU. This choice is made with practical application in mind. The method is based on a lightweight CNN. Based on the face detection, we crop the images with different margins for training and evaluation. 

\begin{figure}
\begin{center}
   \includegraphics[width=1.0\linewidth]{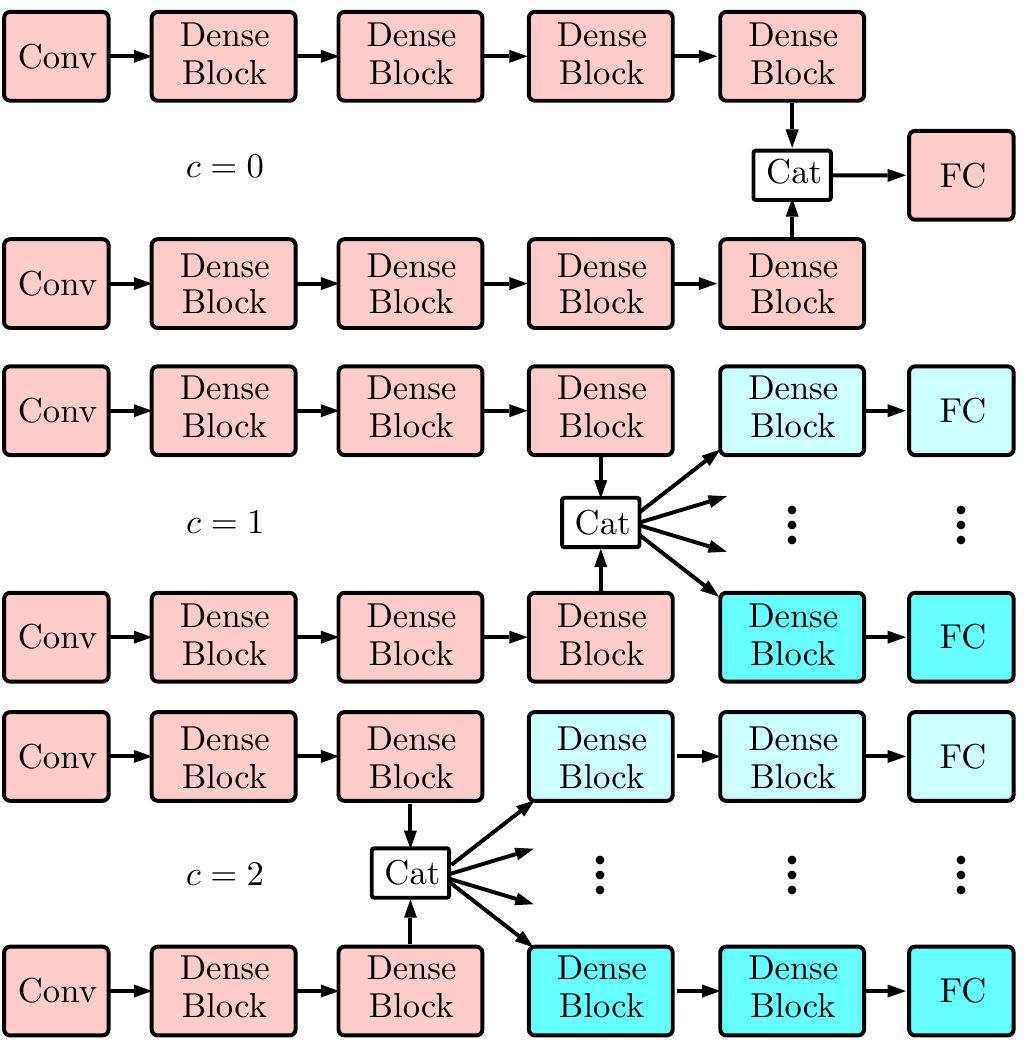}
\end{center}
   \caption{Illustration of the hyperparameter $c$ as the concatenation point. $c=0$ resembles the MTL scenario of hard parameter sharing. Different shades of blue indicate that parameters are not shared between the paths. Red indicates that parameters are shared between the paths. FC denotes the fully-connected output layer.}
\label{fig:cat_points}
\end{figure}

Next, a pair of cropped face images are fed into another CNN. This CNN is inspired by Siamese architectures \cite{yi2014deep} as both images are first processed by individual CNN paths with shared parameters. Then, the features from both paths are concatenated and jointly processed by the rest of the network. After the concatenation point, we split the network up into several convolutional paths with one path for every matching score that is predicted. Finally, the matching scores are predicted at the end of each individual path. This concept follows the soft parameter sharing concept, related to MTL \cite{ruder2017overview}. We also compare this to hard parameter sharing where we use only one path after concatenation. Here, the task is only split at the output layer such that the different matching scores largely share parameters. Thus, we consider the concatenation point $c$ as a hyperparameter which is illustrated in Figure~\ref{fig:cat_points}. For the backbone architecture, we rely on Dense \cite{huang2017densely}. For comparison we also consider a ResNet \cite{he2016deep} basis.

\begin{figure}
\begin{center}
   \includegraphics[width=0.8\linewidth]{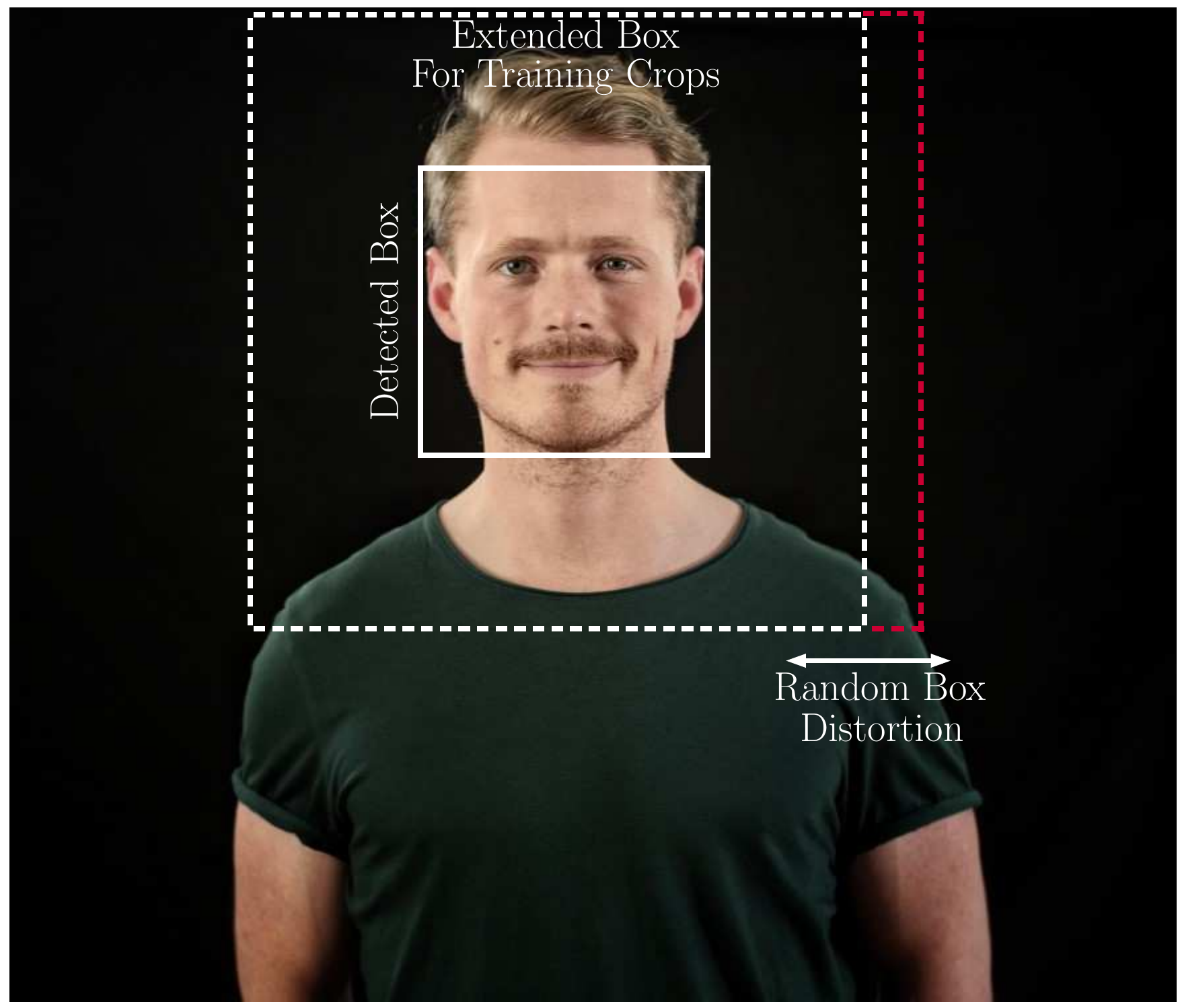}
\end{center}
   \caption{An example of our cropping strategy during training. First, we detect the face using a standard face detection. Second, we extend the box to twice the size, also including some background. Third, we slightly distort the bounding box border. Finally, we take a randomly resized crop from the enlarged bounding box for training.}
\label{fig:cropping}
\end{figure}

During training, we first crop the images around the faces. Here, we add a margin of $m_b = 0.5$ at each side, i.e. the bounding box length and width are doubled. To account for variation in the bounding box detection, we also randomly distort the bounding box borders slightly by up to $\SI{5}{\percent}$ of the box size. Then, we randomly crop from these images, similar to the Inception preprocessing \cite{szegedy2015going}. The process is depicted in Figure~\ref{fig:cropping}. The extended box should increase robustness towards different amounts and types of background being present in the final crops. 

In addition, we apply data augmentation with random brightness, contrast, saturation, and hue changes. Also, we perform horizontal flipping and use dropout with a dropout probability of $p=0.2$. In each epoch, we draw a random match for each user in the training set, restricted to other users in the training set. Training is performed with the Adam algorithm \cite{kingma2014adam} for $300$ epochs. The learning rate and batch size are determined by a small grid search and validation performance. We use the mean squared error between predictions and targets as the loss function:

\begin{equation}
	\mathcal{L} = \frac{1}{N_M}\sum_{j=1}^{N_M}\frac{1}{N_B}\sum_{i=1}^{N_B}(y_j^i-\hat{y}_j^i)^2
\end{equation} 

where $N_B$ is the batch size, $N_M$ is the number of target outputs, $\hat{y}$ is a prediction and $y$ is a ground-truth value. Note that each loss term for each output $M_j$ is only propagated through its designated path. After the concatenation point, the entire loss function affects the network's gradient updates.

To achieve invariance towards switching the input images during evaluation, we perform two forward passes during evaluation. As the computation for the joint processing path is identical, only the computations of the last few layers need to be repeated. The images are cropped to contain the face with a varying cropping margin $m_c$ which determines the zoom on to the faces. We consider different values for $m_c$ in order to study the trade-off between zooming onto the face and also considering a few surroundings.

\section{Results}

\subsection{Evaluation Metrics}

For evaluation, we consider the mean absolute (MAE) as an absolute metric and Pearson's correlation coefficient (PCC) as a relative metric. The MAE is defined as 

\begin{equation}
	\mathit{MAE} = \frac{1}{N}\sum_{i=1}^{N}\mathit{abs}(y_j^i-\hat{y}_j^i)
\end{equation} 

where $N$ is the number of all pairs in the validation or test set.

The PCC is defined as 

\begin{equation}
	\mathit{PCC} = \frac{\sum_{i=1}^{N}(y_j^i-\bar{y}_j)(\hat{y}_j^i-\bar{\hat{y}}_j)}{\sqrt{\sum_{i=1}^{N}(y_j^i-\bar{y}_j)^2}\sqrt{\sum_{i=1}^{N}(\hat{y}_j^i-\bar{\hat{y}}_j)^2}}
\end{equation} 

where $\bar{y_j}$ is the mean predicted value for output $j$ and $\bar{\hat{y}}_j$ is the mean target value for output $j$.

For experiments with a classification task, we use the binary accuracy as a metric.

\subsection{Dataset Characteristics}

First, we put our dataset into context by considering other datasets for tasks of estimating the demographic properties age and sex which are also contained in our dataset. For these two tasks, we use a standard single-path Densenet121 \cite{huang2017densely} which is fed with images from our dataset. We apply the same face cropping and training strategy as for the Siamese Multi-Task CNN. 

Age estimation is typically grouped into biological age estimation and apparent age estimation. In our case, age is self-reported and thus, does not immediately belong to either group but can be interpreted as biological age estimation with a certain amount of noise due to self-reporting. The results for age regression on our dataset compared to other datasets are shown in Table~\ref{tab:age_reg}. All values for the other datasets are based on the recent survey by Carlett et al. \cite{carletti2019age}. We consider the best values that have been reported in the said survey.

\begin{table}
	\centering
	\begin{tabular}{l l l}
	\hline
	Dataset & Dataset Size & MAE \\
	\hline
	MORPH-II \cite{ricanek2006morph} & \num{55608} & \num{2.81} \\
	FG-NET \cite{lanitis2002toward} & \num{1002} & \num{2.00}  \\
	FACES \cite{ebner2010faces} & \num{2052} & \num{3.82}  \\
	LIFESPAN \cite{minear2004lifespan} & \num{1142} & \num{3.87}  \\
	CACD \cite{chen2014cross} & \num{163446} & \num{4.68}  \\
	WebFace \cite{zheng2012visual} & \num{59930} & \num{5.75} \\
	Ours & \num{8300} & \num{4.88} \\
	\hline	
	\end{tabular}
	\caption{Comparison of different datasets for biological age estimation. The best values according to Carlett et al. \cite{carletti2019age} are reported.}
	\label{tab:age_reg}
\end{table}

Overall, we observe that performance on our dataset is slightly lower but similar to the other datasets. This is particularly notable as our dataset is rather small. At the same time, most methods for these datasets rely on pretraining on the IMDB-WIKI dataset \cite{rothe2018deep} with \num{523061} images, providing an additional advantage. Also, age values in our dataset are self-reported which might be associated with some reporting bias. 

Next, we consider the task of sex classification with our dataset. Again, values in our dataset are self-reported. To put our results into context, we consider recent results reported by Haider et al. \cite{haider2019deepgender} and Aslam et al. \cite{aslam2019wavelet}. For the comparison, we consider methods and datasets that used a controlled environment, i.e., mostly frontal photographs of the face. Also, we only consider recent results using deep learning methods. The results are shown in Table~\ref{tab:sex_class}. 

\begin{table}
	\centering
	\begin{tabular}{l l l}
	\hline
	Dataset & Dataset Size & Accuracy($\si{\percent}$) \\
	\hline
	Attributes 25K \cite{zhang2014panda} & \num{24963} & \num{94.10} \\
	Mug shot \cite{juefei2016deepgender} & \num{90000} & \num{97.95} \\
	FERET \cite{phillips1998feret} & \num{5786} & \num{98.84} \\
	SoF \cite{afifi2019afif4} & \num{2662} & \num{98.52} \\
	Ours & \num{8300} & \num{98.30} \\
	\hline	\\
	\end{tabular}
	\caption{Comparison of different methods for sex classification. We consider values reported by Haider et al. \cite{haider2019deepgender} and Aslam et al. \cite{aslam2019wavelet} for datasets with a controlled environment.}
	\label{tab:sex_class}
\end{table}

Again, our results are similar to those reported on other datasets. Thus, overall, our dataset is suitable for estimating both age and sex from face images. This can be considered as a basic quality check which confirms that facial features necessary for estimating demographic properties can be extracted from the images and used by CNNs.

\subsection{Performance and Related Tasks}

Next, we consider the task of estimating the five similarity properties using our Siamese Multi-Task CNN. First, we consider general properties of the similarity scores. Some of the scores are likely not fully independent, e.g., general interests and leisure activities should be related. Table~\ref{tab:m_corr} shows the correlation between all scores in our datasets. 

\begin{table}
\centering
   \begin{tabular}{l | l l l l l}
   & $M_1$ & $M_2$ & $M_3$ & $M_4$ & $M_5$ \\
   \hline
   $M_1$ & 1 & 0.72 & 0.43 & 0.74 & 0.64 \\
   $M_2$ & 0.72 & 1 & 0.48 & 0.83 & 0.74 \\
   $M_3$ & 0.43 & 0.48 & 1 & 0.44 & 0.74 \\
   $M_4$ & 0.74 & 0.83 & 0.44 & 1 & 0.71 \\
   $M_5$ & 0.64 & 0.74 & 0.74 & 0.71 & 1 \\
   \hline  \\
   \end{tabular}
   \caption{Correlation between the different criteria across the entire dataset. The criteria are leisure activities ($M_1$), general interests ($M_2$), relationship preferences ($M_3$), lifestyle ($M_4$) and value system ($M_5$).}
\label{tab:m_corr}
\end{table}

We observe that the different matching criteria are highly correlated. This indicates that agreement or disagreement in one matching score is also associated with said agreement or disagreement in other matching scores. $M_3$ stands out as there is as a lower correlation compared to the correlation between most other scores. However, $M_3$ shows a similar correlation level with $M_5$ (value system). 

\begin{table}
\centering
   \begin{tabular}{l l l}
   & PCC ($\si{\percent}$) & MAE \\
   \hline
   $M_1$ & $26.1 \pm 2.4$ & $0.145 \pm 0.05$  \\
   $M_2$ & $28.3 \pm 3.0$ & $0.158 \pm 0.04$ \\
   $M_3$ & $22.2 \pm 2.2$ & $0.144 \pm 0.04$ \\
   $M_4$ & $30.9 \pm 3.3$ & $0.190 \pm 0.03$ \\
   $M_5$ & $28.0 \pm 2.9$ & $0.150 \pm 0.03$ \\
   $M_{S}$ & $34.7 \pm 3.5$ & $0.173 \pm 0.06$ \\
   \hline  \\
   \end{tabular}
   \caption{Results for a cross-validation experiment for all scores. The standard deviation is calculated over all cross-validation folds.}
\label{tab:m_all}
\end{table}

\begin{table*}
\centering
   \begin{tabular}{l l l l l l l l l l l l l}
   & \multicolumn{2}{c}{$M_1$} & \multicolumn{2}{c}{$M_2$} & \multicolumn{2}{c}{$M_3$} & \multicolumn{2}{c}{$M_4$} & \multicolumn{2}{c}{$M_5$} & \multicolumn{2}{c}{$M_S$} \\
   Configuration & PCC & MAE & PCC & MAE & PCC & MAE & PCC & MAE & PCC & MAE & PCC & MAE  \\
   \hline
   Densenet121 & $26.4$ & $0.145$ & $27.7$ & $\bm{0.160}$ & $22.6$ & $\bm{0.143}$ & $\bm{30.8}$ & $0.190$ & $27.6$ & $0.153$ & $\bm{34.9}$ & $\bm{0.174}$ \\
   Densenet121* & $16.9$ & $0.201$ & $15.1$ & $0.224$ & $11.1$ & $0.191$ & $18.7$ & $0.255$ & $17.1$ & $0.197$ & $21.2$ & $0.238$ \\
   Densenet161 & $24.3$ & $0.150$ & $26.2$ & $0.163$ & $\bm{23.1}$ & $0.144$ & $28.4$ & $0.199$ & $25.3$ & $0.159$ & $32.5$ & $0.183$ \\
   Densenet169 & $\bm{27.1}$ & $\bm{0.142}$ & $27.0$ & $0.161$ & $20.9$ & $0.147$ & $28.1$ & $0.200$ & $26.5$ & $0.156$ & $33.3$ & $0.178$ \\
   ResNet50 & $26.5$ & $0.145$ & $\bm{28.0}$ & $0.161$ & $22.2$ & $0.146$ & $30.4$ & $\bm{0.187}$ & $\bm{28.1}$ & $\bm{0.150}$ & $33.9$ & $0.176$ \\ \hline 
   $c=0$ & $22.4$ & $0.161$ & $22.7$ & $0.187$ & $19.0$ & $0.153$ & $26.2$ & $0.218$ & $23.6$ & $0.166$ & $29.9$ & $0.190$ \\
   $c=2$ & $19.2$ & $0.178$ & $21.1$ & $0.199$ & $16.7$ & $0.163$ & $23.9$ & $0.230$ & $21.5$ & $0.178$ & $26.0$ & $0.216$ \\
   Individual & $18.4$ & $0.182$ & $19.5$ & $0.208$ & $16.0$ & $0.166$ & $20.7$ & $0.241$ & $20.1$ & $0.183$ & $23.4$ & $0.229$ \\
   \hline  \\
   \end{tabular}
   \caption{Results for various ablation experiments. The first block shows results for different baseline architectures. Densenet121* was trained from scratch. The second block shows results for different concatenation points $c$. Individual refers to the training of separate networks. Our standard configuration uses Densenet121 and $c=1$.}
\label{tab:m_all}
\end{table*}

Second, we consider the results for training on our dataset. Table~\ref{tab:m_all} shows the results for three-fold cross-validation. Each fold was defined similarly to our test set. Note that we only evaluate on our test set for all other experiments. The folds are non-overlapping. In general, the correlation coefficient is statistically significant for all $M_i$ ($p < 0.05$) when testing for the null hypothesis that there is no relationship between predictions and targets using Student's t-test on the statistic $t = r*\sqrt{N-2}/\sqrt{1-r^2}$ where $r$ is a correlation coefficient and $N$ is the number of samples. We do observe slight variations between folds but overall the results appear to be consistent. The score $M_3$ stands out with a slightly lower PCC than the other scores. This might be related to the correlation between the scores and $M_3$ which is substantially lower than the correlation between other scores. 

As the task we address is unique so far, we put the performance results into context by considering other related tasks where properties are predicted from faces. For the prediction of age, a more objective measure, we achieve a PCC of $\num{0.850}$ on our dataset. More subjective properties such as facial beauty have been estimated with a correlation of $\num{0.48}$ \cite{rothe2016some} using CNNs and the Gray dataset \cite{gray2010predicting}. For this dataset, facial beauty was defined as the average rating by $30$ users. Also, Rothe et al. tried to predict a user's rating for an image in the context of dating \cite{rothe2016some}. When using visual features only, Rothe et al. achieved a correlation of $\num{0.52}$. For the prediction of personality traits from faces using deep learning, correlation has not been reported. For a randomized training/test split an accuracy of $\SI{67}{\percent}$ was achieved \cite{gavrilescu2017predicting}. Zhang et al. reported the binary accuracy for predicting $20$ personality traits with an average accuracy of $\SI{59.0}{\percent}$ \cite{zhang2017physiognomy}. If we break down our task at hand into binary classification using the median score as the decision threshold, we achieve an accuracy of $\SI{70.3}{\percent}$.

Thus, we can observe overall that our task at hand appears to be more difficult than predicting facial beauty or attractiveness. At the same time, predicting personality seems to be slightly harder than our task of predicting interests, preferences, and attitude. 

\subsection{Ablation Experiments}

Next, we consider several ablation experiments where we vary some parts of our architecture. First, we consider different baseline architectures, see Table~\ref{tab:m_all}. When using a Densenet121 without pretrained weights we observe a substantial decline in performance. This is likely linked to our relatively small dataset size. When using other baseline architectures of similar type with more parameters (Densenet161,Densenet169), performance is similar. This suggests that a larger and deeper model is not necessary for the given task and dataset size. The more different model ResNet50 leads to overall similar performance. Thus, our approach works well with different baseline models. 

Second, we consider architecture variations with our standard configuration. Moving the concatenation point to $c=0$ leads to slightly reduced performance. When moving the concatenation point further to the input with $c=2$ we observe a more substantial decline in performance. This indicates that the different matching scores benefit from a larger, jointly learned representation. Also, there appears to be a sweet spot in terms of the optimal network portion sharing weights. When training individual models for each matching score, we observe a further decline in performance. This underlines the observation that learning the different scores simultaneously benefits performance.

\subsection{Zoom on Faces}

\begin{figure}
\begin{center}
   \includegraphics[width=1.0\linewidth]{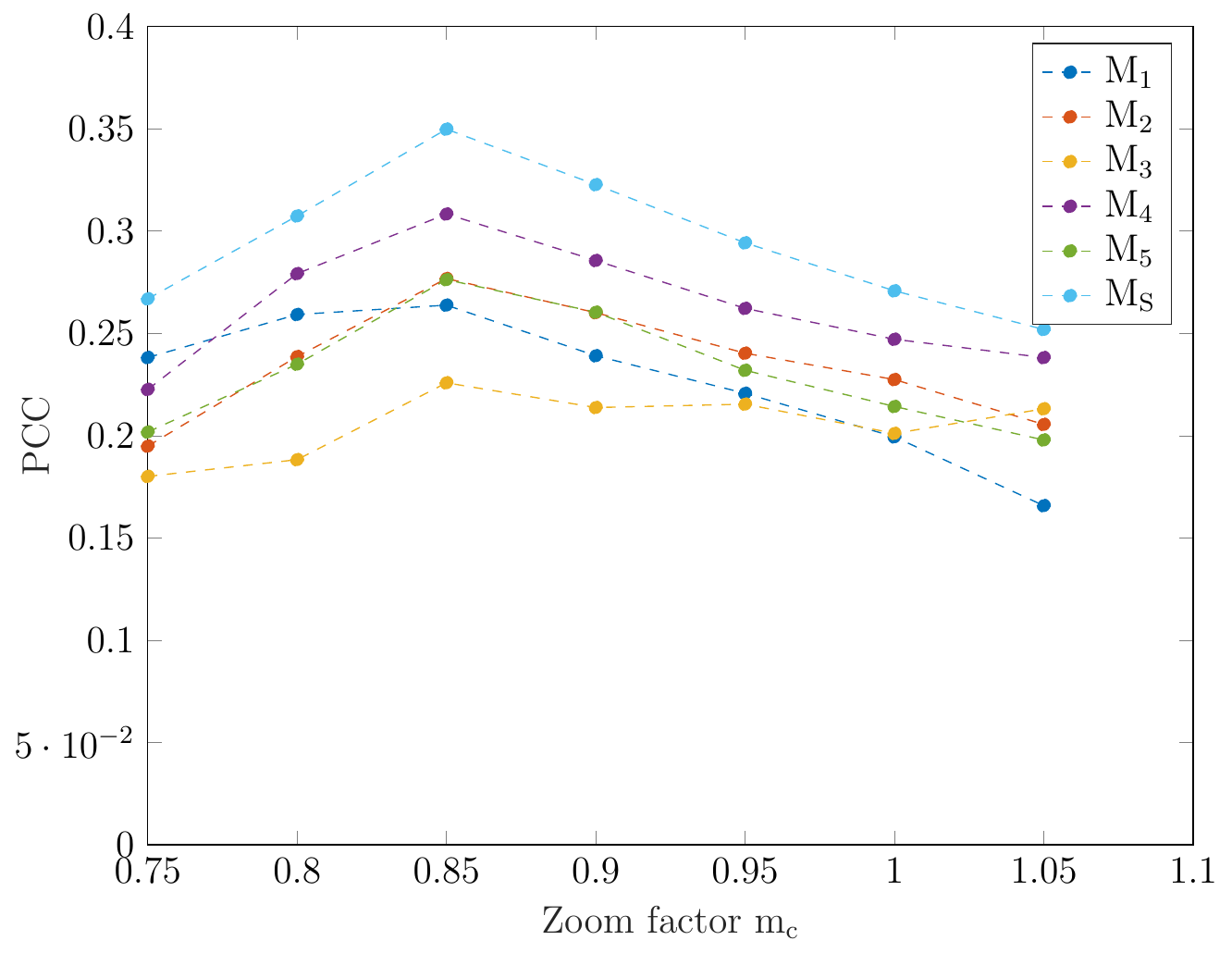}
\end{center}
   \caption{Results for different zoom settings $m_c$ for all matching scores. $m_c < 1$ leads to a stronger zoom on the face. $m_c > 1$ leads to an extension of the bounding box, i.e., zooming out of the face.}
\label{fig:res_zoom}
\end{figure}

Next, we investigate the effect of using different zoom settings onto faces. While a stronger zoom might highlight more detailed features, zooming out might provide additional, potentially relevant features such as hair (style). Therefore, we test different zoom values $m_c$ which controls the zoom into and out of the face. $m_c$ is multiplied with the final bounding boxes' side length. Thus, a zoom of $m_c=1.0$ corresponds to the normal bounding box, $m_c < 1.0$ zooms on to the face and $m_c>1.0$ zooms out of the face. Results for different zoom settings, varied during the testing phase, are shown in Figure~\ref{fig:res_zoom}.

We can observe, that there appears to be a sweet spot, where performance is optimal. Zooming too much onto the face degrades performance. Also, zooming away from the face also reduces performance. Overall, all matching scores change similarly.

\begin{figure*}
\begin{center}
\subfloat{\label{sfig:a}\includegraphics[width=.14\textwidth]{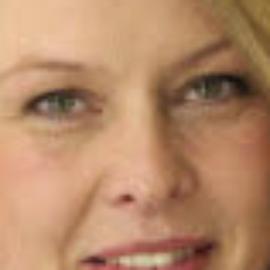}}\hfill
\subfloat{\label{sfig:b}\includegraphics[width=.14\textwidth]{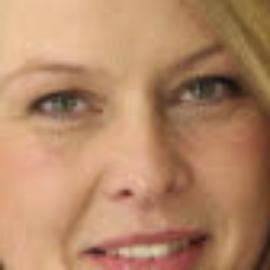}}\hfill
\subfloat{\label{sfig:c}\includegraphics[width=.14\textwidth]{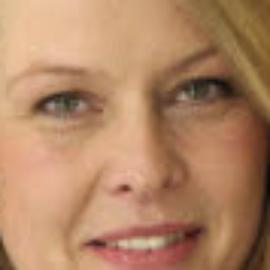}}\hfill
\subfloat{\label{sfig:d}\includegraphics[width=.14\textwidth]{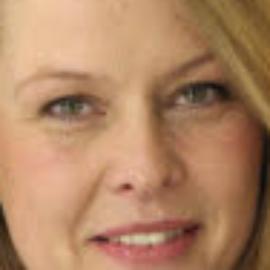}}\hfill
\subfloat{\label{sfig:d1}\includegraphics[width=.14\textwidth]{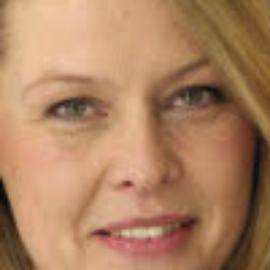}}\hfill
\subfloat{\label{sfig:d2}\includegraphics[width=.14\textwidth]{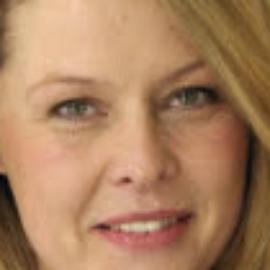}}\hfill
\subfloat{\label{sfig:e}\includegraphics[width=.14\textwidth]{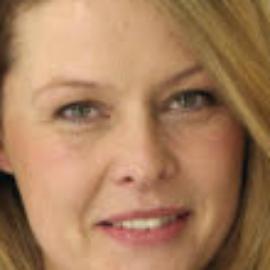}}\\
\subfloat{\label{sfig:f}\includegraphics[width=.14\textwidth]{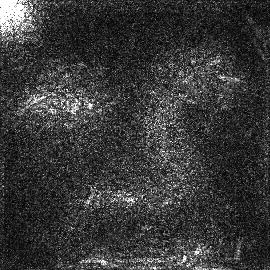}}\hfill
\subfloat{\label{sfig:g}\includegraphics[width=.14\textwidth]{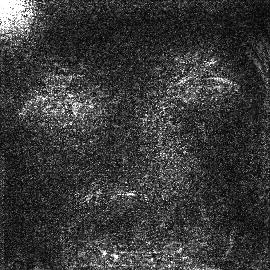}}\hfill
\subfloat{\label{sfig:h}\includegraphics[width=.14\textwidth]{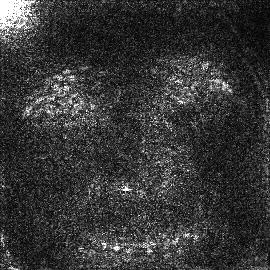}}\hfill
\subfloat{\label{sfig:i}\includegraphics[width=.14\textwidth]{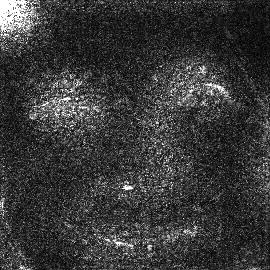}}\hfill
\subfloat{\label{sfig:i1}\includegraphics[width=.14\textwidth]{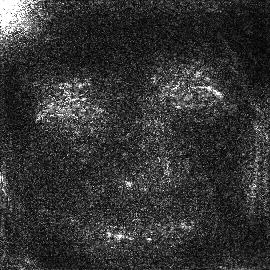}}\hfill
\subfloat{\label{sfig:i2}\includegraphics[width=.14\textwidth]{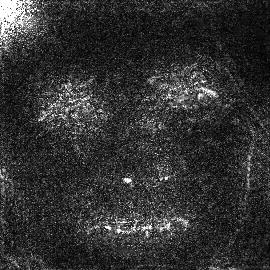}}\hfill
\subfloat{\label{sfig:j}\includegraphics[width=.14\textwidth]{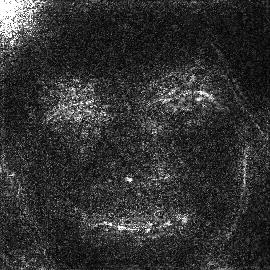}}\\

\subfloat{\label{sfig:a}\includegraphics[width=.14\textwidth]{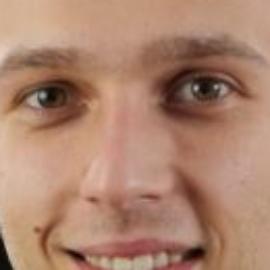}}\hfill
\subfloat{\label{sfig:b}\includegraphics[width=.14\textwidth]{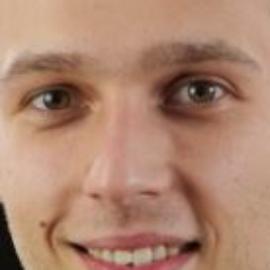}}\hfill
\subfloat{\label{sfig:c}\includegraphics[width=.14\textwidth]{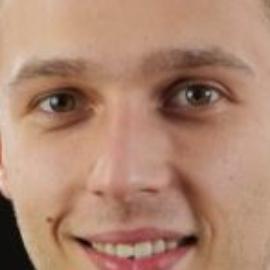}}\hfill
\subfloat{\label{sfig:d}\includegraphics[width=.14\textwidth]{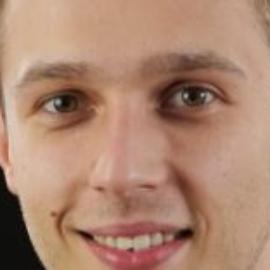}}\hfill
\subfloat{\label{sfig:d1}\includegraphics[width=.14\textwidth]{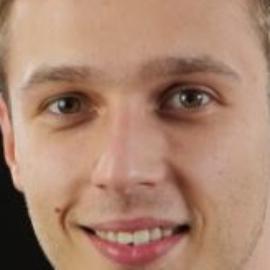}}\hfill
\subfloat{\label{sfig:d2}\includegraphics[width=.14\textwidth]{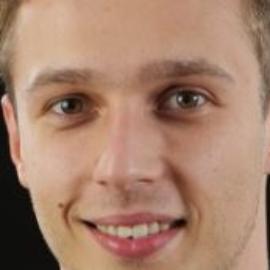}}\hfill
\subfloat{\label{sfig:e}\includegraphics[width=.14\textwidth]{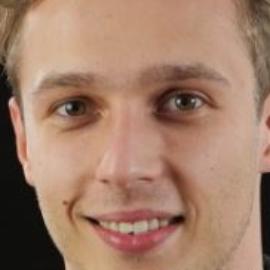}}\\
\subfloat{\label{sfig:f}\includegraphics[width=.14\textwidth]{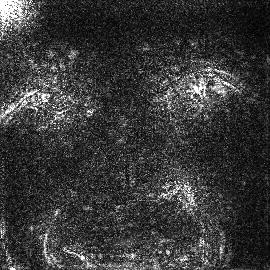}}\hfill
\subfloat{\label{sfig:g}\includegraphics[width=.14\textwidth]{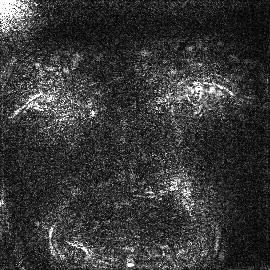}}\hfill
\subfloat{\label{sfig:h}\includegraphics[width=.14\textwidth]{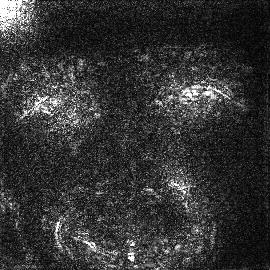}}\hfill
\subfloat{\label{sfig:i}\includegraphics[width=.14\textwidth]{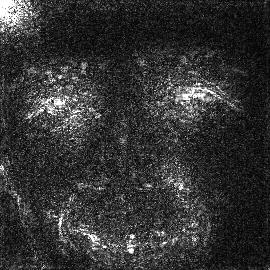}}\hfill
\subfloat{\label{sfig:i1}\includegraphics[width=.14\textwidth]{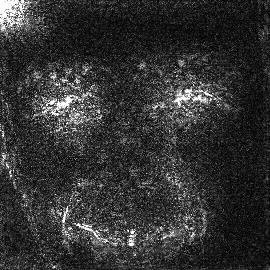}}\hfill
\subfloat{\label{sfig:i2}\includegraphics[width=.14\textwidth]{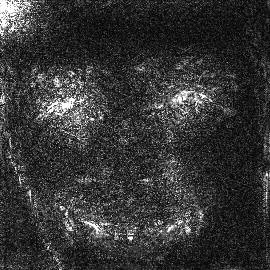}}\hfill
\subfloat{\label{sfig:j}\includegraphics[width=.14\textwidth]{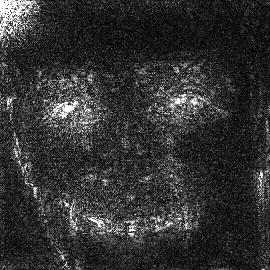}}\\
\end{center}
   \caption{Example images with different values $m_c$ and the corresponding saliency maps.}
\label{fig:res_saliency}
\end{figure*}

\begin{table*}
\centering
   \begin{tabular}{l l l l l l l l l l l l l}
   & \multicolumn{2}{c}{$M_1$} & \multicolumn{2}{c}{$M_2$} & \multicolumn{2}{c}{$M_3$} & \multicolumn{2}{c}{$M_4$} & \multicolumn{2}{c}{$M_5$} & \multicolumn{2}{c}{$M_S$} \\
   Configuration & PCC & MAE & PCC & MAE & PCC & MAE & PCC & MAE & PCC & MAE & PCC & MAE  \\
   \hline
   Standard & $26.4$ & $0.145$ & $27.7$ & $0.160$ & $22.6$ & $0.143$ & $30.8$ & $0.190$ & $27.6$ & $0.153$ & $34.9$ & $0.174$ \\
   $m_c = 1.5$ & $14.6$ & $0.211$ & $17.8$ & $0.195$ & $19.1$ & $0.152$ & $20.3$ & $0.218$ & $17.9$ & $0.188$ & $22.2$ & $0.219$ \\
   $m_c = 2.0$ & $13.0$ & $0.195$ & $16.7$ & $0.199$ & $16.3$ & $0.168$ & $19.9$ & $0.210$ & $15.1$ & $0.204$ & $20.5$ & $0.226$ \\
   No Boxes & $9.26$ & $0.237$ & $12.7$ & $0.233$ & $6.32$ & $0.221$ & $13.5$ & $0.273$ & $12.4$ & $0.219$ & $14.1$ & $0.263$ \\
   \hline  \\
   \end{tabular}
   \caption{Results for background inclusion. For our standard configuration, we use a zoom of $mc=0.85$. No Boxes refers to training with full images without prior face detection.}
\label{tab:background}
\end{table*}

To investigate this aspect further, we also considered saliency maps for visualization of the relevant features for the task at hand, see Figure~\ref{fig:res_saliency}. We employ guided backpropagation \cite{springenberg2014striving}, enhanced by SmoothGrad \cite{smilkov2017smoothgrad} with \num{100} repetitions, Gaussian noise and a noise level $\sigma$ depending on the image's intensity range \cite{uozbulak_pytorch_vis_2019}. Across all zooms, we can observe that the model appears to recognize multiple facial features. In particular, there seems to be a focus on regions around the eyes and the mouth. When zooming on to the face, the region around the mouth appears to get partially cropped out, potentially relating to the drop off in performance. Similarly, when zooming out of the face, the face borders become more visible, potentially opening up space for irrelevant or misleading features.

Besides small zoom changes on the face, including the entire background could also be of interest. A lot of images contain background that might also provide information for the matching scores. At the same time, the background could also be misleading, e.g., if a nonathletic person provides a photograph taken during a sports activity. In addition, when choosing larger cropping areas, we trade off detailed face information for more context which might also affect performance. Therefore, we study the effect of providing background to our model, see Table~\ref{tab:background}. First, we consider zooming away further from the face with $m_c = 1.5$ and $m_c = 2.0$. Also, we train a model with full images and no face cropping. There is a decline in performance for further zooming which is even worse for training on entire images. This suggests that using a face detection first and zooming onto the face is beneficial for our problem. Of course, we cannot rule out that background can also help for this task but it might require a different approach, e.g., using multiple input paths with different resolutions. 

Overall, the choice of zoom appears to have a large impact on performance, both for small zoom changes around the face and more extensive zoom settings, away from the face. Saliency maps reveal that certain facial features appear to be relevant for the model, visible across multiple zoom settings.

%\subsection{Intramatch Variation}

\subsection{Self-Matching}

\begin{table}
\centering
   \begin{tabular}{l l l l l l l l}
   \multicolumn{2}{l}{Config.} & $M_1$ & $M_2$ & $M_3$ & $M_4$ & $M_5$ & $M_S$ \\
   \hline
   $\SI{0}{\percent}$ & MS & $0.15$ & $0.16$ & $0.14$ & $0.19$ & $0.15$ & $0.17$  \\
   $\SI{10}{\percent}$ & MS & $0.25$ & $0.18$ & $0.15$ & $0.19$ & $0.16$ & $0.20$  \\ \hline
   $\SI{0}{\percent}$ & IM & $0.25$ & $0.23$ & $0.20$ & $0.15$ & $0.17$ & $0.15$  \\  
   $\SI{10}{\percent}$ & IM & $0.01$ & $0.03$ & $0.09$ & $0.03$ & $0.02$ & $0.01$  \\ \hline   
   $\SI{0}{\percent}$ & SM & $0.26$ & $0.26$ & $0.23$ & $0.18$ & $0.20$ & $0.19$  \\
   $\SI{10}{\percent}$ & SM & $0.03$ & $0.18$ & $0.17$ & $0.15$ & $0.12$ & $0.07$  \\   
   \hline  \\
   \end{tabular}
   \caption{MAE for evaluation of identity matching (IM) and self matching (SM) in comparison to our normal matching scores (MS). In our standard configuration, we do not train on self or identity matches ($\SI{0}{\percent}$). Note that the PCC is not meaningful for this experiment as all targets have a value of $1$ for the IM and SM task.}
\label{tab:selfmatch}
\end{table}

Another interesting aspect of our approach is the problem of self matching. Based on the design of the matching score, a person matching with himself should have a perfect score of one, as all of his interests match with himself. This can be understood as an implicit face recognition problem that naturally arises with this task. During training, we do not explicitly enforce this property as we do not train on self matches. However, it appears reasonable that our model could learn this form of implicit face recognition if explicitly trained for it. Therefore, we conducted an experiment where we fine-tuned our model additionally where $\SI{10}{\percent}$ of all training examples are self or identity matches, i.e., either the identical image or another image of the same user. 

We perform three evaluations for this strategy. First, we assess the normal matching score prediction performance as it might degrade due to the additional fine-tuning. Second, we assess the performance for identity matches, i.e., matching with the exact same image. Third, we evaluate the self matching performance where another image of the user is used for evaluation. This resembles the implicit face recognition task. The results for this experiment are shown in Table~\ref{tab:selfmatch}. For our standard model that was not trained for the task, we observe an MAE that is slightly worse than the normal matching performance for the self matching task. For the identity task, performance is slightly better than for the normal matching task. When training both an self and identity matches, the identity task is solved almost perfectly and for the self matching task, performance is improved substantially. 

\begin{table}
\centering
   \begin{tabular}{l l l l l l l}
   Config. & $M_1$ & $M_2$ & $M_3$ & $M_4$ & $M_5$ & $M_S$ \\
   \hline
   2 Pairs & $0.06$ & $0.09$ & $0.06$ & $0.10$ & $0.08$ & $0.09$  \\
   3 Pairs & $0.07$ & $0.10$ & $0.06$ & $0.11$ & $0.08$ & $0.10$  \\ 
   4 Pairs & $0.07$ & $0.10$ & $0.06$ & $0.11$ & $0.08$ & $0.10$  \\  
   2 Pairs all & $0.06$ & $0.08$ & $0.05$ & $0.08$ & $0.06$ & $0.08$  \\  
   3 Pairs all & $0.05$ & $0.07$ & $0.05$ & $0.07$ & $0.06$ & $0.07$  \\
   4 Pairs all & $0.06$ & $0.08$ & $0.06$ & $0.09$ & $0.07$ & $0.09$  \\   
   \hline  \\
   \end{tabular}
   \caption{Mean absolute prediction difference between image pairs of the same two people, averaged across the entire test set. We differentiate by the number of image pairs that are available for each match. Also, we consider both distinct image pairs, where both images must differ and all image pairs, where every image of user 1 is matched with every image of user 2.}
\label{tab:variation}
\end{table}

The observation that our model does not naturally learn face recognition well also implies that predictions might differ for the same two people and two different images. We investigate this hypothesis by considering the mean absolute difference between predictions obtained from different image pairs for the same two people. The results are shown in Table~\ref{tab:variation}. We can observe that there is a certain variation between predictions for different image pairs. In particular, it is notable that for completely different image pairs only, the mean difference increases. Thus, overall, the model's limited face recognition capability is also reflected in match variation.

\section{Limitations}

We find several interesting properties that come with our task and we observe a significant correlation between predicted and target matching scores, indicating that properties related to interests, preferences, and attitude can be partially derived from face images. Still, there are several limitations that need to be discussed. First, our dataset is relatively small and captures a certain demographic and cultural background. Thus, we cannot rule out effects that are specific to culture and generalization might be limited. Furthermore, it should be kept in mind that our targets are based on self-reported properties. Thus, we cannot rule out that our model also learned some underlying reporting bias that is not immediately obvious. 

Also, we observed a certain variation for predictions with different images of the same two people which raises the question of consistency. The model's predictions might also depend on situational properties such as facial expression. Considering face interpretation in social sciences, several studies have observed that people's interpretation of faces is easily influenced by facial expression or pose \cite{todorov2008understanding,naumann2009personality,zebrowitz2011ecological,todorov2015social}. This could indicate that the matching scores are not necessarily or at least not only related to facial features but also current facial expression. This relates to recent findings that the type of images users uploaded to social networks is significantly influenced by the users personality \cite{liu2016analyzing}. Thus, there might also be an implicit relationship between the self-reported values in the psychological test and the \textit{type} of images users uploaded. While this would open up our approach for manipulation, in case the effect is known to the user, the underlying hypothesis of scores being learnable from face images would still be valid. It does, however, open up interesting future research questions. For example, the matching score's predictability could be investigated if face images are controlled in terms of facial expression or users are instructed to provide a certain type of images.

\section{Conclusion}

In this paper, we investigate the feasibility of predicting multiple matching scores, related to interests, preferences, and attitude, from the face images of two people. The goal is to predict potential compatibility between the two people for starting a romantic relationship. The pairwise matching scores between the two people were obtained with a psychological test designed for capturing interests, preferences, and attitude. We try to learn the matching scores in a supervised manner with a Siamese Multi-Task CNN that initially processes the two face images with two paths sharing parameters, followed by individual prediction paths. The approach works consistently with different backbone architectures and we observe that partial parameter sharing is beneficial for predicting the correlated targets. We find that the zoom onto faces affects performance and that including a lot of background is not beneficial for our approach. Given the design of our matching problem, we also observe that face recognition can be learned implicitly with our approach if we train on self matches. While the overall predictive performance is limited, we find a significant correlation between predictions and targets. Thus, some facial features or facial expressions appear to be loosely related to our matching scores. For future work, more informative features could be provided to the model, e.g., in the form of multiple images or video material. Also, the effect of users' image choice could be investigated further.

{\small
\bibliographystyle{ieee_fullname}
\bibliography{egbib}

\begin{thebibliography}{10}\itemsep=-1pt

\bibitem{afifi2019afif4}
Mahmoud Afifi and Abdelrahman Abdelhamed.
\newblock Afif4: deep gender classification based on adaboost-based fusion of
  isolated facial features and foggy faces.
\newblock {\em Journal of Visual Communication and Image Representation},
  62:77--86, 2019.

\bibitem{akehurst2011ccr}
Joshua Akehurst, Irena Koprinska, Kalina Yacef, Luiz Pizzato, Judy Kay, and
  Tomasz Rej.
\newblock Ccr—a content-collaborative reciprocal recommender for online
  dating.
\newblock In {\em Twenty-Second International Joint Conference on Artificial
  Intelligence}, 2011.

\bibitem{albright1988consensus}
Linda Albright, David~A Kenny, and Thomas~E Malloy.
\newblock Consensus in personality judgments at zero acquaintance.
\newblock {\em Journal of personality and social psychology}, 55(3):387, 1988.

\bibitem{altwaijry2013relative}
Hani Altwaijry and Serge Belongie.
\newblock Relative ranking of facial attractiveness.
\newblock In {\em 2013 IEEE Workshop on Applications of Computer Vision
  (WACV)}, pages 117--124. IEEE, 2013.

\bibitem{aslam2019wavelet}
Aasma Aslam, Khizar Hayat, Arif~Iqbal Umar, Bahman Zohuri, Payman Zarkesh-Ha,
  David Modissette, Sahib~Zar Khan, and Babar Hussian.
\newblock Wavelet-based convolutional neural networks for gender
  classification.
\newblock {\em Journal of Electronic Imaging}, 28(1):013012, 2019.

\bibitem{bak2015gast}
Peter~Michael Bak.
\newblock {\em Zu Gast in Deiner Wirklichkeit: Empathie als Schl{\"u}ssel
  gelungener Kommunikation}.
\newblock Springer-Verlag, 2015.

\bibitem{barr2018detecting}
Makenzie Barr, Guodong Guo, Sarah Colby, and Melissa Olfert.
\newblock Detecting body mass index from a facial photograph in lifestyle
  intervention.
\newblock {\em Technologies}, 6(3):83, 2018.

\bibitem{brozovsky2007recommender}
Lukas Brozovsky and Vaclav Petricek.
\newblock Recommender system for online dating service.
\newblock {\em arXiv preprint cs/0703042}, 2007.

\bibitem{cai2010learning}
Xiongcai Cai, Michael Bain, Alfred Krzywicki, Wayne Wobcke, Yang~Sok Kim, Paul
  Compton, and Ashesh Mahidadia.
\newblock Learning collaborative filtering and its application to people to
  people recommendation in social networks.
\newblock In {\em 2010 IEEE International Conference on Data Mining}, pages
  743--748. IEEE, 2010.

\bibitem{campbell2016initial}
Lorne Campbell, Kristi Chin, and Sarah~CE Stanton.
\newblock Initial evidence that individuals form new relationships with
  partners that more closely match their ideal preferences.
\newblock {\em Collabra: Psychology}, 2(1), 2016.

\bibitem{carletti2019age}
Vincenzo Carletti, Antonio Greco, Gennaro Percannella, and Mario Vento.
\newblock Age from faces in the deep learning revolution.
\newblock {\em IEEE transactions on pattern analysis and machine intelligence},
  2019.

\bibitem{caruana1997multitask}
Rich Caruana.
\newblock Multitask learning.
\newblock {\em Machine learning}, 28(1):41--75, 1997.

\bibitem{chen2014cross}
Bor-Chun Chen, Chu-Song Chen, and Winston~H Hsu.
\newblock Cross-age reference coding for age-invariant face recognition and
  retrieval.
\newblock In {\em European conference on computer vision}, pages 768--783.
  Springer, 2014.

\bibitem{chung2017two}
Dahjung Chung, Khalid Tahboub, and Edward~J Delp.
\newblock A two stream siamese convolutional neural network for person
  re-identification.
\newblock In {\em Proceedings of the IEEE International Conference on Computer
  Vision}, pages 1983--1991, 2017.

\bibitem{clemens2015influence}
Chris Clemens, David Atkin, and Archana Krishnan.
\newblock The influence of biological and personality traits on gratifications
  obtained through online dating websites.
\newblock {\em Computers in Human Behavior}, 49:120--129, 2015.

\bibitem{ebner2010faces}
Natalie~C Ebner, Michaela Riediger, and Ulman Lindenberger.
\newblock Faces—a database of facial expressions in young, middle-aged, and
  older women and men: Development and validation.
\newblock {\em Behavior research methods}, 42(1):351--362, 2010.

\bibitem{eisenthal2006facial}
Yael Eisenthal, Gideon Dror, and Eytan Ruppin.
\newblock Facial attractiveness: Beauty and the machine.
\newblock {\em Neural Computation}, 18(1):119--142, 2006.

\bibitem{finkel2012online}
Eli~J Finkel, Paul~W Eastwick, Benjamin~R Karney, Harry~T Reis, and Susan
  Sprecher.
\newblock Online dating: A critical analysis from the perspective of
  psychological science.
\newblock {\em Psychological Science in the Public interest}, 13(1):3--66,
  2012.

\bibitem{fiore2010s}
Andrew~T Fiore, Lindsay~Shaw Taylor, Xiaomeng Zhong, Gerald~A Mendelsohn, and
  Coye Cheshire.
\newblock Who's right and who writes: People, profiles, contacts, and replies
  in online dating.
\newblock In {\em 2010 43rd Hawaii International Conference on System
  Sciences}, pages 1--10. IEEE, 2010.

\bibitem{fu2010age}
Yun Fu, Guodong Guo, and Thomas~S Huang.
\newblock Age synthesis and estimation via faces: A survey.
\newblock {\em IEEE transactions on pattern analysis and machine intelligence},
  32(11):1955--1976, 2010.

\bibitem{gan2014deep}
Junying Gan, Lichen Li, Yikui Zhai, and Yinhua Liu.
\newblock Deep self-taught learning for facial beauty prediction.
\newblock {\em Neurocomputing}, 144:295--303, 2014.

\bibitem{gao2018automatic}
Lian Gao, Weixin Li, Zehua Huang, Di Huang, and Yunhong Wang.
\newblock Automatic facial attractiveness prediction by deep multi-task
  learning.
\newblock In {\em 2018 24th International Conference on Pattern Recognition
  (ICPR)}, pages 3592--3597. IEEE, 2018.

\bibitem{gavrilescu2017predicting}
Mihai Gavrilescu and Nicolae Vizireanu.
\newblock Predicting the sixteen personality factors (16pf) of an individual by
  analyzing facial features.
\newblock {\em EURASIP Journal on Image and Video Processing}, 2017(1):59,
  2017.

\bibitem{girshick2015fast}
Ross Girshick.
\newblock Fast r-cnn.
\newblock In {\em Proceedings of the IEEE international conference on computer
  vision}, pages 1440--1448, 2015.

\bibitem{gray2010predicting}
Douglas Gray, Kai Yu, Wei Xu, and Yihong Gong.
\newblock Predicting facial beauty without landmarks.
\newblock In {\em European Conference on Computer Vision}, pages 434--447.
  Springer, 2010.

\bibitem{gurovich2019identifying}
Yaron Gurovich, Yair Hanani, Omri Bar, Guy Nadav, Nicole Fleischer, Dekel
  Gelbman, Lina Basel-Salmon, Peter~M Krawitz, Susanne~B Kamphausen, Martin
  Zenker, et~al.
\newblock Identifying facial phenotypes of genetic disorders using deep
  learning.
\newblock {\em Nature medicine}, 25(1):60, 2019.

\bibitem{hadsell2006dimensionality}
Raia Hadsell, Sumit Chopra, and Yann LeCun.
\newblock Dimensionality reduction by learning an invariant mapping.
\newblock In {\em 2006 IEEE Computer Society Conference on Computer Vision and
  Pattern Recognition (CVPR'06)}, volume~2, pages 1735--1742. IEEE, 2006.

\bibitem{haider2019deepgender}
Khurram~Zeeshan Haider, Kaleem~Razzaq Malik, Shehzad Khalid, Tabassam Nawaz,
  and Sohail Jabbar.
\newblock Deepgender: real-time gender classification using deep learning for
  smartphones.
\newblock {\em Journal of Real-Time Image Processing}, 16(1):15--29, 2019.

\bibitem{hassin2000facing}
Ran Hassin and Yaacov Trope.
\newblock Facing faces: studies on the cognitive aspects of physiognomy.
\newblock {\em Journal of personality and social psychology}, 78(5):837, 2000.

\bibitem{he2016deep}
Kaiming He, Xiangyu Zhang, Shaoqing Ren, and Jian Sun.
\newblock Deep residual learning for image recognition.
\newblock In {\em Proceedings of the IEEE conference on computer vision and
  pattern recognition}, pages 770--778, 2016.

\bibitem{huang2017densely}
Gao Huang, Zhuang Liu, Laurens Van Der~Maaten, and Kilian~Q Weinberger.
\newblock Densely connected convolutional networks.
\newblock In {\em Proceedings of the IEEE conference on computer vision and
  pattern recognition}, pages 4700--4708, 2017.

\bibitem{juefei2016deepgender}
Felix Juefei-Xu, Eshan Verma, Parag Goel, Anisha Cherodian, and Marios
  Savvides.
\newblock Deepgender: Occlusion and low resolution robust facial gender
  classification via progressively trained convolutional neural networks with
  attention.
\newblock In {\em Proceedings of the IEEE conference on computer vision and
  pattern recognition workshops}, pages 68--77, 2016.

\bibitem{kingma2014adam}
Diederik~P Kingma and Jimmy Ba.
\newblock Adam: A method for stochastic optimization.
\newblock {\em arXiv preprint arXiv:1412.6980}, 2014.

\bibitem{krizhevsky2012imagenet}
Alex Krizhevsky, Ilya Sutskever, and Geoffrey~E Hinton.
\newblock Imagenet classification with deep convolutional neural networks.
\newblock In {\em Advances in neural information processing systems}, pages
  1097--1105, 2012.

\bibitem{krzywicki2010interaction}
Alfred Krzywicki, Wayne Wobcke, Xiongcai Cai, Ashesh Mahidadia, Michael Bain,
  Paul Compton, and Yang~Sok Kim.
\newblock Interaction-based collaborative filtering methods for recommendation
  in online dating.
\newblock In {\em International Conference on Web Information Systems
  Engineering}, pages 342--356. Springer, 2010.

\bibitem{kutty2014people}
Sangeetha Kutty, Richi Nayak, and Lin Chen.
\newblock A people-to-people matching system using graph mining techniques.
\newblock {\em World Wide Web}, 17(3):311--349, 2014.

\bibitem{lanitis2002toward}
Andreas Lanitis, Christopher~J. Taylor, and Timothy~F Cootes.
\newblock Toward automatic simulation of aging effects on face images.
\newblock {\em IEEE Transactions on pattern Analysis and machine Intelligence},
  24(4):442--455, 2002.

\bibitem{levi2015age}
Gil Levi and Tal Hassner.
\newblock Age and gender classification using convolutional neural networks.
\newblock In {\em Proceedings of the iEEE conference on computer vision and
  pattern recognition workshops}, pages 34--42, 2015.

\bibitem{li2019incorporating}
Zhihong Li, Yining Song, and Xiaoying Xu.
\newblock Incorporating facial attractiveness in photos for online dating
  recommendation.
\newblock {\em Electronic Commerce Research}, 19(2):285--310, 2019.

\bibitem{liang2018scut}
Lingyu Liang, Luojun Lin, Lianwen Jin, Duorui Xie, and Mengru Li.
\newblock Scut-fbp5500: A diverse benchmark dataset for multi-paradigm facial
  beauty prediction.
\newblock In {\em 2018 24th International Conference on Pattern Recognition
  (ICPR)}, pages 1598--1603. IEEE, 2018.

\bibitem{liu2016analyzing}
Leqi Liu, Daniel Preotiuc-Pietro, Zahra~Riahi Samani, Mohsen~E Moghaddam, and
  Lyle Ungar.
\newblock Analyzing personality through social media profile picture choice.
\newblock In {\em Tenth international AAAI conference on web and social media},
  2016.

\bibitem{mcnulty2016should}
James~K McNulty.
\newblock Should spouses be demanding less from marriage? a contextual
  perspective on the implications of interpersonal standards.
\newblock {\em Personality and social psychology bulletin}, 42(4):444--457,
  2016.

\bibitem{minear2004lifespan}
Meredith Minear and Denise~C Park.
\newblock A lifespan database of adult facial stimuli.
\newblock {\em Behavior Research Methods, Instruments, \& Computers},
  36(4):630--633, 2004.

\bibitem{misra2016cross}
Ishan Misra, Abhinav Shrivastava, Abhinav Gupta, and Martial Hebert.
\newblock Cross-stitch networks for multi-task learning.
\newblock In {\em Proceedings of the IEEE Conference on Computer Vision and
  Pattern Recognition}, pages 3994--4003, 2016.

\bibitem{naumann2009personality}
Laura~P Naumann, Simine Vazire, Peter~J Rentfrow, and Samuel~D Gosling.
\newblock Personality judgments based on physical appearance.
\newblock {\em Personality and social psychology bulletin}, 35(12):1661--1671,
  2009.

\bibitem{ng2012recognizing}
Choon~Boon Ng, Yong~Haur Tay, and Bok-Min Goi.
\newblock Recognizing human gender in computer vision: a survey.
\newblock In {\em Pacific Rim International Conference on Artificial
  Intelligence}, pages 335--346. Springer, 2012.

\bibitem{niu2016ordinal}
Zhenxing Niu, Mo Zhou, Le Wang, Xinbo Gao, and Gang Hua.
\newblock Ordinal regression with multiple output cnn for age estimation.
\newblock In {\em Proceedings of the IEEE conference on computer vision and
  pattern recognition}, pages 4920--4928, 2016.

\bibitem{uozbulak_pytorch_vis_2019}
Utku Ozbulak.
\newblock Pytorch cnn visualizations.
\newblock \url{https://github.com/utkuozbulak/pytorch-cnn-visualizations},
  2019.

\bibitem{penton2006personality}
Ian~S Penton-Voak, Nicholas Pound, Anthony~C Little, and David~I Perrett.
\newblock Personality judgments from natural and composite facial images: More
  evidence for a “kernel of truth” in social perception.
\newblock {\em Social cognition}, 24(5):607--640, 2006.

\bibitem{phillips1998feret}
P~Jonathon Phillips, Harry Wechsler, Jeffery Huang, and Patrick~J Rauss.
\newblock The feret database and evaluation procedure for face-recognition
  algorithms.
\newblock {\em Image and vision computing}, 16(5):295--306, 1998.

\bibitem{qi2016sketch}
Yonggang Qi, Yi-Zhe Song, Honggang Zhang, and Jun Liu.
\newblock Sketch-based image retrieval via siamese convolutional neural
  network.
\newblock In {\em 2016 IEEE International Conference on Image Processing
  (ICIP)}, pages 2460--2464. IEEE, 2016.

\bibitem{ranjan2017hyperface}
Rajeev Ranjan, Vishal~M Patel, and Rama Chellappa.
\newblock Hyperface: A deep multi-task learning framework for face detection,
  landmark localization, pose estimation, and gender recognition.
\newblock {\em IEEE Transactions on Pattern Analysis and Machine Intelligence},
  41(1):121--135, 2017.

\bibitem{ricanek2006morph}
Karl Ricanek and Tamirat Tesafaye.
\newblock Morph: A longitudinal image database of normal adult age-progression.
\newblock In {\em 7th International Conference on Automatic Face and Gesture
  Recognition (FGR06)}, pages 341--345. IEEE, 2006.

\bibitem{rothe2016some}
Rasmus Rothe, Radu Timofte, and Luc Van~Gool.
\newblock Some like it hot-visual guidance for preference prediction.
\newblock In {\em Proceedings of the IEEE conference on computer vision and
  pattern recognition}, pages 5553--5561, 2016.

\bibitem{rothe2018deep}
Rasmus Rothe, Radu Timofte, and Luc Van~Gool.
\newblock Deep expectation of real and apparent age from a single image without
  facial landmarks.
\newblock {\em International Journal of Computer Vision}, 126(2-4):144--157,
  2018.

\bibitem{ruder2017overview}
Sebastian Ruder.
\newblock An overview of multi-task learning in deep neural networks.
\newblock {\em arXiv preprint arXiv:1706.05098}.

\bibitem{smilkov2017smoothgrad}
Daniel Smilkov, Nikhil Thorat, Been Kim, Fernanda Vi{\'e}gas, and Martin
  Wattenberg.
\newblock Smoothgrad: removing noise by adding noise.
\newblock {\em arXiv preprint arXiv:1706.03825}, 2017.

\bibitem{smith2013online}
Aaron~Whitman Smith and Maeve Duggan.
\newblock {\em Online dating \& relationship}.
\newblock Pew Research Center Washington, DC, 2013.

\bibitem{springenberg2014striving}
Jost~Tobias Springenberg, Alexey Dosovitskiy, Thomas Brox, and Martin
  Riedmiller.
\newblock Striving for simplicity: The all convolutional net.
\newblock {\em arXiv preprint arXiv:1412.6806}, 2014.

\bibitem{szegedy2015going}
Christian Szegedy, Wei Liu, Yangqing Jia, Pierre Sermanet, Scott Reed, Dragomir
  Anguelov, Dumitru Erhan, Vincent Vanhoucke, and Andrew Rabinovich.
\newblock Going deeper with convolutions.
\newblock In {\em Proceedings of the IEEE conference on computer vision and
  pattern recognition}, pages 1--9, 2015.

\bibitem{todorov2015social}
Alexander Todorov, Christopher~Y Olivola, Ron Dotsch, and Peter
  Mende-Siedlecki.
\newblock Social attributions from faces: Determinants, consequences, accuracy,
  and functional significance.
\newblock {\em Annual review of psychology}, 66:519--545, 2015.

\bibitem{todorov2008understanding}
Alexander Todorov, Chris~P Said, Andrew~D Engell, and Nikolaas~N Oosterhof.
\newblock Understanding evaluation of faces on social dimensions.
\newblock {\em Trends in cognitive sciences}, 12(12):455--460, 2008.

\bibitem{tu2014online}
Kun Tu, Bruno Ribeiro, David Jensen, Don Towsley, Benyuan Liu, Hua Jiang, and
  Xiaodong Wang.
\newblock Online dating recommendations: matching markets and learning
  preferences.
\newblock In {\em Proceedings of the 23rd international conference on world
  wide web}, pages 787--792. ACM, 2014.

\bibitem{watson1989strangers}
David Watson.
\newblock Strangers' ratings of the five robust personality factors: Evidence
  of a surprising convergence with self-report.
\newblock {\em Journal of Personality and Social Psychology}, 57(1):120, 1989.

\bibitem{whyte2017things}
Stephen Whyte and Benno Torgler.
\newblock Things change with age: Educational assortment in online dating.
\newblock {\em Personality and Individual Differences}, 109:5--11, 2017.

\bibitem{wilf2015method}
Itzhak Wilf, Yael Shor, Shai Gilboa, David Gavriel, and Gilad Bechar.
\newblock Method and system for predicting personality traits, capabilities and
  suggested interactions from images of a person, 2015.
\newblock US Patent App. 14/700,315.

\bibitem{willis2006first}
Janine Willis and Alexander Todorov.
\newblock First impressions: Making up your mind after a 100-ms exposure to a
  face.
\newblock {\em Psychological science}, 17(7):592--598, 2006.

\bibitem{xia2016design}
Peng Xia, Shuangfei Zhai, Benyuan Liu, Yizhou Sun, and Cindy Chen.
\newblock Design of reciprocal recommendation systems for online dating.
\newblock {\em Social Network Analysis and Mining}, 6(1):32, 2016.

\bibitem{xu2017facial}
Jie Xu, Lianwen Jin, Lingyu Liang, Ziyong Feng, Duorui Xie, and Huiyun Mao.
\newblock Facial attractiveness prediction using psychologically inspired
  convolutional neural network (pi-cnn).
\newblock In {\em 2017 IEEE International Conference on Acoustics, Speech and
  Signal Processing (ICASSP)}, pages 1657--1661. IEEE, 2017.

\bibitem{yan2014cost}
Haibin Yan.
\newblock Cost-sensitive ordinal regression for fully automatic facial beauty
  assessment.
\newblock {\em Neurocomputing}, 129:334--342, 2014.

\bibitem{yi2014deep}
Dong Yi, Zhen Lei, Shengcai Liao, and Stan~Z Li.
\newblock Deep metric learning for person re-identification.
\newblock In {\em 2014 22nd International Conference on Pattern Recognition},
  pages 34--39. IEEE, 2014.

\bibitem{zebrowitz2011ecological}
Leslie~A Zebrowitz.
\newblock Ecological and social approaches to face perception.
\newblock {\em The Oxford handbook of face perception}, pages 31--50, 2011.

\bibitem{zebrowitz1997accurate}
Leslie~A Zebrowitz and Mary~Ann Collins.
\newblock Accurate social perception at zero acquaintance: The affordances of a
  gibsonian approach.
\newblock {\em Personality and social psychology review}, 1(3):204--223, 1997.

\bibitem{zhang2014panda}
Ning Zhang, Manohar Paluri, Marc'Aurelio Ranzato, Trevor Darrell, and Lubomir
  Bourdev.
\newblock Panda: Pose aligned networks for deep attribute modeling.
\newblock In {\em Proceedings of the IEEE conference on computer vision and
  pattern recognition}, pages 1637--1644, 2014.

\bibitem{zhang2017faceboxes}
Shifeng Zhang, Xiangyu Zhu, Zhen Lei, Hailin Shi, Xiaobo Wang, and Stan~Z Li.
\newblock Faceboxes: A cpu real-time face detector with high accuracy.
\newblock In {\em 2017 IEEE International Joint Conference on Biometrics
  (IJCB)}, pages 1--9. IEEE, 2017.

\bibitem{zhang2017physiognomy}
Ting Zhang, Ri-Zhen Qin, Qiu-Lei Dong, Wei Gao, Hua-Rong Xu, and Zhan-Yi Hu.
\newblock Physiognomy: Personality traits prediction by learning.
\newblock {\em International Journal of Automation and Computing},
  14(4):386--395, 2017.

\bibitem{zhang2014facial}
Zhanpeng Zhang, Ping Luo, Chen~Change Loy, and Xiaoou Tang.
\newblock Facial landmark detection by deep multi-task learning.
\newblock In {\em European conference on computer vision}, pages 94--108.
  Springer, 2014.

\bibitem{zheng2012visual}
Song Zheng.
\newblock {\em Visual image recognition system with object-level image
  representation}.
\newblock PhD thesis, 2012.

\end{thebibliography}
}

\end{document}